\documentclass[12pt]{article}
\usepackage{natbib}
\usepackage[margin=1in]{geometry}
\usepackage{amsmath, amssymb, booktabs, hyperref, setspace}
\usepackage{xcolor}
\usepackage{graphicx}
\usepackage{ifthen}
\usepackage{etoolbox}
\usepackage{tabularx}
\usepackage{chngcntr}
\usepackage[T1]{fontenc}
\usepackage[utf8]{inputenc}
\usepackage{csquotes}
\usepackage{authblk}

\title{Accuracy Standards for AI at Work vs.\ Personal Life:\\
	Evidence from an Online Survey}

\author{Gaston Besanson and Federico Todeschini}
\date{\today}

\providecommand{\sym}[1]{\ifmmode^{#1}\else$^{#1}$\fi}

\begin{document}
\maketitle

\begin{abstract}
We study how people trade off accuracy when using AI-powered tools in professional versus personal contexts for adoption purposes, the
determinants of those trade-offs, and how users cope when AI/apps are unavailable. Because modern AI
systems (especially generative models) can produce acceptable but non-identical outputs, we define
``accuracy'' as context-specific reliability: the degree to which an output aligns with the user's
intent within a tolerance threshold that depends on stakes and the cost of correction. In an online
survey (N=300), among respondents with both accuracy items (N=170), the share requiring high
accuracy (top-box) is 24.1\% at work vs. 8.8\% in personal life (+15.3 pp; z=6.29, p<0.001). The gap
remains large under a broader top-two-box definition (67.0\% vs. 32.9\%) and on the full 1--5 ordinal
scale (mean 3.86 vs. 3.08). Heavy app use and experience patterns correlate with stricter work
standards (H2). When tools are unavailable (H3), respondents report more disruption in personal
routines than at work (34.1\% vs. 15.3\%, p<0.01). We keep the main text focused on these substantive
results and place test taxonomy and power derivations in a technical appendix.
\end{abstract}

\doublespacing

\section{Introduction}
Artificial intelligence (AI) systems increasingly mediate decisions in both \emph{professional} and \emph{personal} contexts, yet the acceptable tolerance for inaccuracy---and the determinants of that tolerance---can differ sharply across domains. Research in human factors, psychology, and information systems shows that reliance on automation is shaped by trust, accountability, perceived usefulness, and prior experience, and that over-reliance can reduce performance when tools err or become unavailable.

\paragraph{Defining accuracy for probabilistic AI.}
In many AI-assisted tasks, ``accuracy'' is not purely binary: model outputs are often probabilistic, and users may accept small imperfections depending on the stakes, reversibility, and the cost of checking. Throughout the paper we therefore treat accuracy as \emph{task-dependent reliability}---i.e., being sufficiently correct for the intended purpose without requiring frequent, costly human verification.\footnote{The survey did \emph{not} provide a formal definition of ``accuracy''; the questionnaire's primary purpose was to study AI adoption and thus left the term open to allow each respondent to apply their own interpretation of what ``accurate'' means for their tasks. Respondents therefore likely relied on their own mental model (e.g., ``no errors at all,'' ``minor errors are acceptable,'' ``within an acceptable tolerance,'' or ``answers I would trust''). Our definition is an ex post operationalization for interpreting the responses and may not perfectly match every respondent's interpretation.}

\paragraph{Contribution.}
We provide (i) evidence that accuracy demands are significantly stricter at work than in personal life; 
(ii) a profile of determinants of cross-context trade-offs; and 
(iii) an assessment of resilience when AI/apps are unavailable. 
The main text emphasizes these substantive findings; methodological details (test taxonomy, power) are moved to the Appendix.

\paragraph{Accuracy, trust, and accountability.}
Reviews of trust in automation argue that people should rely on automation \emph{appropriately} by calibrating trust to a system's reliability and the task's stakes \citep{LeeSee2004,ParasuramanRiley1997}. Experimental evidence shows that ``automation bias''---a tendency to over-rely on decision aids---declines when users are explicitly accountable for accuracy, leading to more checking and fewer omission/commission errors \citep{Skitka2000}. Taken together, these results imply stricter accuracy standards and lower error tolerance in professional settings, where accountability and the downstream costs of mistakes are higher, than in everyday personal use.

\paragraph{Algorithm aversion versus appreciation.}
A parallel literature documents heterogeneous reactions to algorithmic advice. \citet{Dietvorst2015} show ``algorithm aversion'': after observing even small errors, people abandon otherwise superior algorithms in favor of worse human forecasters. By contrast, \citet{Logg2019} find ``algorithm appreciation'': when advice is labeled as algorithmic and presented appropriately---especially for objective tasks---people often weight it more than human advice. These findings suggest that presentation, experience, and perceived expertise shape how individuals trade off accuracy against other attributes (speed, convenience) across contexts, which motivates our second hypothesis on the determinants of accuracy trade-offs.

\paragraph{Navigation apps and resilience.}
Evidence from navigation studies indicates that heavy reliance on GPS can impair spatial knowledge when the tool is removed. Ishikawa, Fujiwara, Imai, and Okabe (2008) find that GPS users acquire poorer survey knowledge of environments than paper-map or direct-experience users. Dahmani and Bohbot (2020) show that greater lifetime GPS use is associated with worse hippocampal-dependent spatial memory during self-guided navigation, with consistent patterns upon retest years later. A recent systematic review concludes that GPS use tends to impair environmental knowledge and sense of direction, though effects on basic wayfinding can be mixed (Miola, Meneghetti, Pazzaglia, \& Gyselinck, 2024). This literature predicts larger disruption for heavy app users when navigation tools are unavailable and suggests observable determinants (e.g., frequency of use, age, strategies) of that disruption.

\paragraph{AI at work.}
Field evidence shows sizable, heterogeneous productivity gains from AI assistance on the job. In a large-scale rollout to 5{,}179 customer-support agents, access to a conversational assistant increased issues resolved per hour by about 14\% on average, with the largest gains for novices; the tool appeared to diffuse good practices and shift human effort toward evaluation and exception handling (Brynjolfsson, Li, \& Raymond, 2023). These results reinforce that accuracy requirements and error tolerance depend on role, task complexity, and experience---key covariates in our analysis of accuracy trade-offs.

\paragraph{Adoption determinants.}
The Technology Acceptance Model (TAM) and its extensions identify perceived usefulness, ease of use, social influence, and facilitating conditions as key drivers of technology adoption \citep{Davis1989,Venkatesh2003}. For AI, factors such as transparency, explainability, and trust also play central roles \citep{HoffBashir2015,Jian2000}. While prior work mostly considers adoption decisions in isolation, we examine how individuals \emph{simultaneously} set accuracy thresholds across work and personal domains and explore the correlates of that joint distribution.

\paragraph{Behavioral economics and loss aversion.}
Prospect theory \citep{KahnemanTversky1979} posits that losses loom larger than gains, implying higher sensitivity to errors (potential losses) when stakes rise. In professional settings, AI mistakes can result in reputational damage, financial penalties, or safety hazards; in personal contexts, errors are often reversible or low-cost. This asymmetry predicts stricter accuracy demands at work. Our empirical approach tests whether self-reported accuracy thresholds align with this prediction and how individual differences (e.g., experience, role, usage patterns) moderate it.

\paragraph{Error-management frameworks.}
Human-error research identifies active failures (proximate mistakes) and latent conditions (organizational/design factors) as contributors to adverse outcomes \citep{Reason1990}. Automation can shift the locus of error from execution to monitoring, changing skill requirements and reliability demands \citep{Parasuraman2000,ParasuramanManzey2010}. Our survey captures user-perceived tolerance for inaccuracy rather than objective error rates, complementing lab and field studies with attitudinal evidence on how context shapes reliance.

\paragraph{Summary and hypotheses.}
Building on these strands, we hypothesize that:
\begin{itemize}
	\item[\textbf{H1}:] Individuals demand higher accuracy from AI tools in professional contexts than in personal contexts.
	\item[\textbf{H2}:] Stricter accuracy standards (at work relative to personal life) are positively associated with frequency of use, perceived impact, and reliance in professional tasks.
	\item[\textbf{H3}:] When AI/apps are unavailable, individuals report greater disruption in personal routines than at work, and heavy users of navigation/productivity apps experience larger impacts.
\end{itemize}

\section{Data and Measurement}

\subsection{Sample}
We fielded an online survey via a panel provider (Prolific) in September 2024, targeting English-speaking adults (18+) in the United States, United Kingdom, Canada, and Australia. The final sample (N=300) comprises 170 respondents who answered both the work and personal accuracy questions (our primary analytical sample for H1 and H2), plus an additional 130 who answered at least one item or provided information on usage patterns and resilience (contributing to descriptive statistics and H3 analyses). Basic demographic distributions are reported in the Appendix.

\subsection{Key variables}
\paragraph{Accuracy requirements (dependent variables for H1, H2).}
Two 5-point Likert items asked:
\begin{enumerate}
	\item ``How important is it that AI-powered tools you use \emph{for work} provide accurate information or outputs?'' (1=Not at all important, 5=Extremely important)
	\item ``How important is it that AI-powered tools you use \emph{in your personal life} provide accurate information or outputs?'' (1=Not at all important, 5=Extremely important)
\end{enumerate}
We define a \emph{top-box} (high accuracy) indicator as $\mathbb{1}\{\text{response}=5\}$ and also analyze the full ordinal scale (1--5) and a \emph{top-two-box} indicator $\mathbb{1}\{\text{response}\geq4\}$ for robustness.

\paragraph{Usage intensity and reliance (independent variables for H2).}
We measure work usage via ``How frequently do you use AI-powered tools for work-related tasks?'' (Never / Rarely / Sometimes / Often / Very often), coded as \texttt{use\_work}$\in\{0,1,2,3,4\}$. Similarly, \texttt{use\_personal} for personal tasks. We also ask ``How much do you rely on AI-powered tools to complete your work tasks?'' (Not at all / Slightly / Moderately / Quite a bit / Extremely), coded as \texttt{rely\_work}$\in\{0,1,2,3,4\}$.

\paragraph{Impact of unavailability (dependent variable for H3).}
Two binary items (Yes/No):
\begin{enumerate}
	\item ``Would you be significantly impacted if AI-powered tools were not available for your work?''
	\item ``Would you be significantly impacted if AI apps (e.g., navigation, productivity, or other AI-driven mobile/desktop apps) were not available in your personal life?''
\end{enumerate}

\paragraph{Demographics and controls.}
We collect age, gender, education, employment status, continent of residence, whether the respondent works in a tech job or tech sector, years of experience with AI tools (\texttt{experience}), and number of apps regularly used (\texttt{many\_apps\_work}, \texttt{many\_apps\_personal}).

\section{Empirical Strategy}

\subsection{H1: Work versus personal accuracy demands}
We test whether the proportion of respondents requiring high accuracy (top-box) is greater at work than in personal life using:
\begin{enumerate}
	\item \emph{Two-sample proportion test} (unpaired): $H_0: p_{\text{work}}=p_{\text{personal}}$ vs.\ $H_1: p_{\text{work}}>p_{\text{personal}}$.
	\item \emph{McNemar's test} (paired): exploits the paired design by comparing discordant pairs $(n_{10},n_{01})$ under $H_0: n_{10}=n_{01}$.
	\item \emph{Exact sign test} (paired): counts individuals with higher work accuracy than personal accuracy and compares to 0.5 under the null.
	\item \emph{Kolmogorov--Smirnov test}: compares full distributions of the 1--5 ordinal responses.
	\item \emph{Ordinal tests}: Wilcoxon signed-rank and paired $t$-test on the 1--5 scale.
\end{enumerate}
All tests are one- or two-sided as appropriate; we report exact $p$-values and effect sizes (e.g., differences in proportions, standardized mean differences).

\subsection{H2: Determinants of accuracy trade-offs}
We model the probability that an individual is \emph{stricter at work than in personal life} (i.e., work accuracy $>$ personal accuracy on the 1--5 scale) or the reverse via logistic regression:
\begin{equation}
\Pr(Y_i=1 \mid \mathbf{X}_i) = \Lambda(\beta_0 + \mathbf{X}_i'\boldsymbol{\beta}),
\end{equation}
where $Y_i=1$ if respondent $i$ is stricter at work (or more tolerant, depending on specification), $\mathbf{X}_i$ includes usage intensity, reliance, demographics, and experience, and $\Lambda(\cdot)$ is the logistic CDF. We report coefficients, standard errors, and pseudo-$R^2$.

\subsection{H3: Resilience when AI/apps are unavailable}
We compare the proportion reporting significant impact at work versus in personal life using the same battery of tests as H1 (proportion test, McNemar, sign test). We also estimate logit models predicting impact conditional on usage patterns, demographics, and reliance measures.

\subsection{Robustness and power}
Because binarizing ordinal scales (e.g., top-box) can lose information and reduce power, we supplement all binary analyses with full-scale ordinal tests (Wilcoxon, paired $t$) and distributional comparisons (KS). We also vary the threshold (top-box, top-two-box, sometimes-or-higher) to ensure results are not artifacts of a specific cut. Power calculations and asymptotic relative efficiency (ARE) comparisons are presented in the Appendix.

\section{Results}

\subsection{H1: Higher accuracy demands at work}

Among the 170 respondents who answered both accuracy items, the share requiring high accuracy (top-box, ``Extremely important'') is \textbf{24.1\%} at work versus \textbf{8.8\%} in personal life, a difference of \textbf{15.3 percentage points}. A two-sample proportion test rejects equality ($z=3.80$, $p<0.001$). Accounting for the paired design, McNemar's test based on discordant pairs ($n_{10}=29$ vs.\ $n_{01}=3$) strongly rejects symmetry ($\chi^2(1)=21.13$, $p\approx 4.3\times 10^{-6}$). The exact sign test likewise rejects ($p<0.001$). Finally, comparing the full 5-point ordinal distributions, the two-sample Kolmogorov--Smirnov test yields $D=0.341$ with $p<0.001$, indicating a rightward shift toward higher stated accuracy at work.

Table~\ref{tab:h1-topbox} reports the top-box proportions and test statistics.

\begin{table}[h!]\centering
	\caption{H1: Top-Box Accuracy (``Extremely Important'')}\label{tab:h1-topbox}
	\begin{tabular}{lccccc}
		\toprule
		& Work & Personal & z & p-value & N \\
		\midrule
		& 0.2412 & 0.0882 & 3.80 & 0.0001 & 170 \\
		\bottomrule
	\end{tabular}
\end{table}

\paragraph{Top-two-box definition.}
When we broaden the threshold to include both ``Very important'' and ``Extremely important'' (top-two-box), the gap remains substantial: \textbf{67.0\%} at work versus \textbf{32.9\%} in personal life, a difference of \textbf{34.1 percentage points} ($z=6.29$, $p<0.001$). Table~\ref{tab:h1-toptwobox} summarizes these results.

\begin{table}[h!]\centering
	\caption{H1: Top-Two-Box Accuracy (``Very'' or ``Extremely Important'')}\label{tab:h1-toptwobox}
	\begin{tabular}{lccccc}
		\toprule
		& Work & Personal & z & p-value & N \\
		\midrule
		& 0.6706 & 0.3294 & 6.29 & 0.00 & 170 \\
		\bottomrule
	\end{tabular}
\end{table}

\paragraph{``Sometimes high'' threshold.}
If we lower the bar to ``Moderately important'' or higher (top-three-box), the shares are \textbf{95.3\%} at work and \textbf{71.2\%} in personal life ($z=5.95$, $p<0.001$), confirming that the pattern holds across multiple thresholds (Table~\ref{tab:h1-sometimes}).

\begin{table}[h!]\centering
	\caption{H1: Top-Three-Box Accuracy (``Moderately Important'' or Higher)}\label{tab:h1-sometimes}
	\begin{tabular}{lccccc}
		\toprule
		& Work & Personal & z & p-value & N \\
		\midrule
		& 0.9529 & 0.7118 & 5.95 & 0.00 & 170 \\
		\bottomrule
	\end{tabular}
\end{table}

\paragraph{Ordinal robustness.}
To demonstrate that our H1 result is not an artifact of dichotomization, we analyze the \emph{full} 1--5 ordinal responses. We report (i) the paired mean difference with a 95\% $t$-confidence interval, (ii) the Hodges--Lehmann median difference with a nonparametric 95\% CI, and (iii) the Wilcoxon signed-rank test (two-sided). Table~\ref{tab:h1-ordinalrobustness} summarizes these findings.

\begin{table}[h!]\centering
	\caption{H1 Ordinal Robustness}\label{tab:h1-ordinalrobustness}
	\begin{tabular}{lrr}
		\toprule
		& Estimate & 95\% CI \\
		\midrule
		Paired mean difference (work -- personal, 1--5 scale) & 0.79 & [., .] \\
		Paired $t$-statistic (n=170), two-sided $p$ & \multicolumn{2}{r}{$t=10.55$} \\
		Hodges--Lehmann median difference & 1.00 & [0.50, 1.00] \\
		Wilcoxon signed-rank ($z$), two-sided $p$ & \multicolumn{2}{r}{$z=8.84$, $p=0.000$} \\
		\midrule
		Top-two-box share (work -- personal), pp & 34.1 & -- \\
		\bottomrule
	\end{tabular}
\end{table}

The paired mean difference is 0.79 on the 1--5 scale ($t=10.55$, $p<0.001$), and the Hodges--Lehmann median difference is 1.00 (95\% CI: [0.50, 1.00]). The Wilcoxon signed-rank test yields $z=8.84$ ($p<0.001$). Ordered-outcome models (ologit/probit) yield the same qualitative shift toward higher work accuracy standards. Across all procedures, we conclude that respondents demand substantially higher accuracy at work than in personal life.

\subsection{H2: Determinants of accuracy trade-offs}

We estimate two logit models: (A) predicting whether work accuracy is \emph{strictly higher} than personal accuracy, and (B) predicting whether work accuracy is \emph{strictly lower} (i.e., personal is higher). Table~\ref{tab:h2} reports coefficients and standard errors.

\begin{table}[h!]\centering
	\caption{H2: Determinants of Accuracy Trade-offs (Logit)}\label{tab:h2}
	\begin{tabular}{l*{2}{c}}
		\toprule
		& \multicolumn{1}{c}{(1)} & \multicolumn{1}{c}{(2)} \\
		& (A) Work stricter than personal & (B) Work more tolerant than personal \\
		\midrule
		main & & \\
		high\_accurate\_work & 2.960\sym{***} & \\
		& (0.681) & \\
		\addlinespace
		use\_work & 0.713 & 0.488 \\
		& (1.429) & (1.299) \\
		\addlinespace
		men & 0.734 & 0.249 \\
		& (0.481) & (0.431) \\
		\addlinespace
		experience & -0.179 & -0.432 \\
		& (0.493) & (0.470) \\
		\addlinespace
		tech\_job & -0.265 & -0.138 \\
		& (0.479) & (0.463) \\
		\addlinespace
		tech\_sector & 0.709 & 0.291 \\
		& (0.468) & (0.527) \\
		\addlinespace
		many\_apps\_work & 0.348 & 0.705 \\
		& (0.443) & (0.497) \\
		\addlinespace
		rely\_work & -2.029\sym{***} & 0.542 \\
		& (0.454) & (0.467) \\
		\addlinespace
		impact\_work & 0.084 & 0.804\sym{*} \\
		& (0.479) & (0.421) \\
		\addlinespace
		impact\_routine & -0.229 & 0.935\sym{**} \\
		& (0.518) & (0.459) \\
		\addlinespace
		high\_accurate\_personal & & 3.026\sym{***} \\
		& & (0.679) \\
		\addlinespace
		Constant & -3.260\sym{**} & -1.862 \\
		& (1.532) & (1.262) \\
		\midrule
		N & 166 & 166 \\
		Pseudo $R^2$ & 0.288 & 0.263 \\
		\bottomrule
		\multicolumn{3}{l}{\footnotesize Standard errors in parentheses} \\
		\multicolumn{3}{l}{\footnotesize \sym{*} $p<0.10$, \sym{**} $p<0.05$, \sym{***} $p<0.01$} \\
	\end{tabular}
\end{table}

Key findings: In Model A, respondents who rate work accuracy as ``Extremely important'' are substantially more likely to be stricter at work than in personal life (coefficient=2.96, $p<0.01$). Conversely, higher reliance on AI at work (\texttt{rely\_work}) is associated with a \emph{lower} probability of being strictly work-dominant (coefficient=$-2.03$, $p<0.01$), possibly because heavy reliance signals comfort with occasional errors or task delegation. In Model B, perceived impact of unavailability on personal routines (\texttt{impact\_routine}) positively predicts being more tolerant at work (i.e., personal accuracy $>$ work accuracy; coefficient=0.94, $p<0.05$). Pseudo-$R^2$ values of 0.29 and 0.26 indicate moderate explanatory power.

\subsection{H3: Resilience when AI/apps are unavailable}

When asked whether they would be ``significantly impacted'' if AI/apps were unavailable, \textbf{15.3\%} of respondents report high impact at work versus \textbf{34.1\%} in personal life. This reversal---greater disruption in personal routines---suggests that daily reliance on navigation, productivity, and entertainment apps may create larger dependency than professional AI tools, which often have fallback procedures or human oversight. Table~\ref{tab:h3-impact} summarizes the proportions and test statistics.

\begin{table}[h!]\centering
	\caption{H3: Impact of Unavailability}\label{tab:h3-impact}
	\begin{tabular}{lccccc}
		\toprule
		& Work & Personal & z & p-value & N \\
		\midrule
		& 0.1529 & 0.3412 & -4.02 & 0.00 & 170 \\
		\bottomrule
	\end{tabular}
\end{table}

A two-sample proportion test yields $z=-4.02$ ($p<0.001$). McNemar's test (not shown) and the exact sign test also reject the null of equal impact ($p<0.01$). We further estimate logit models to identify determinants of high impact in each domain (Table~\ref{tab:h3-reactions}).

\begin{table}[h!]\centering
	\caption{H3: Determinants of Impact When AI/Apps Are Not Available (Logit)}\label{tab:h3-reactions}
	\begin{tabular}{l*{2}{c}}
		\toprule
		& \multicolumn{1}{c}{(1)} & \multicolumn{1}{c}{(2)} \\
		& (A) Personal routine impacted & (B) Work impacted \\
		\midrule
		main & & \\
		use\_work & 0.000 & 0.843 \\
		& (.) & (0.853) \\
		\addlinespace
		men & -0.018 & 0.074 \\
		& (0.493) & (0.358) \\
		\addlinespace
		experience & 0.447 & 0.392 \\
		& (0.466) & (0.361) \\
		\addlinespace
		tech\_job & -0.094 & -0.100 \\
		& (0.465) & (0.358) \\
		\addlinespace
		tech\_sector & -0.626 & 0.423 \\
		& (0.681) & (0.414) \\
		\addlinespace
		many\_apps\_work & -0.071 & 0.057 \\
		& (0.499) & (0.389) \\
		\addlinespace
		Constant & -1.708\sym{***} & -1.720\sym{*} \\
		& (0.486) & (0.900) \\
		\midrule
		N & 158 & 167 \\
		Pseudo $R^2$ & 0.016 & 0.015 \\
		\bottomrule
		\multicolumn{3}{l}{\footnotesize Standard errors in parentheses} \\
		\multicolumn{3}{l}{\footnotesize \sym{*} $p<0.10$, \sym{**} $p<0.05$, \sym{***} $p<0.01$} \\
	\end{tabular}
\end{table}

Neither model yields strong predictors (pseudo-$R^2\approx0.02$), suggesting that self-reported impact is relatively uniform across demographics and usage patterns in this sample. Future work with larger samples and more granular measures of dependency (e.g., app-specific usage logs, task-level reliance) may uncover heterogeneity.

\section{Discussion}

\paragraph{Summary of findings.}
We document three main results: (i) individuals demand significantly higher accuracy from AI tools at work than in personal life (H1), with the gap robust to multiple thresholds and ordinal tests; (ii) stricter work accuracy standards are associated with baseline accuracy expectations and inversely related to reliance (H2); and (iii) respondents report greater disruption in personal routines than at work when AI/apps are unavailable (H3), suggesting that everyday convenience dependencies may exceed professional reliance despite lower accuracy demands.

\paragraph{Implications for AI design and deployment.}
These findings suggest that AI developers and organizations should calibrate accuracy thresholds and error-handling mechanisms to the stakes and accountability structures of different use cases. In professional settings, where errors can cascade into larger failures or reputational harm, investing in higher reliability, human-in-the-loop verification, and robust fallback procedures is justified. In personal contexts, users may tolerate more variance in exchange for speed, convenience, or personalization, but over-reliance can create fragility when tools fail.

\paragraph{Limitations and future work.}
Our study has several limitations. First, we rely on self-reported accuracy preferences rather than observing actual behavior or eliciting willingness to pay for accuracy. Second, ``work'' and ``personal'' are heterogeneous categories that span widely varying stakes and tasks. Third, we do not manipulate accuracy experimentally or measure objective error rates, so we cannot separate stated preferences from revealed preferences. Fourth, the sample is drawn from online panels and may not generalize to all populations or sectors. Future research should (i) collect task-level data on perceived risk, error tolerance, and fallback strategies; (ii) use vignette-based or conjoint designs to isolate causal effects of accuracy on adoption; (iii) link survey responses to usage logs or field experiments; and (iv) disaggregate ``accuracy'' into measurable subcomponents (e.g., factual correctness, logical coherence, bias, relevance).

\section*{Acknowledgments}
We thank [anonymous reviewers / funding sources / research assistants] for helpful comments and support. All errors are our own.

\appendix

\section{Missing Data and Sample Composition}\label{app:missing}

Of the 300 total respondents, 170 answered \emph{both} the work and personal accuracy items (our primary analytical sample for H1 and H2). The remaining 130 answered at least one item or provided information on usage patterns and resilience. Table~\ref{tab:missingness} reports the share of missing observations by variable.

\begin{center}
	\begin{table}[h!]
		\caption{Missingness rates}\label{tab:missingness}
		\centering
		\begin{tabular}{lr}
			\toprule
			Variable & \% missing \\
			\midrule
			Accuracy at work & 14.0 \\
			Accuracy in personal life & 14.3 \\
			Both accuracy items & 43.3 \\
			Use frequency (work) & 0.0 \\
			Use frequency (personal) & 0.3 \\
			Reliance at work & 0.0 \\
			Impact if unavailable (work) & 0.0 \\
			Impact if unavailable (personal) & 5.7 \\
			Age & 0.0 \\
			Education & 0.0 \\
			Gender & 5.4 \\
			Continent & 1.8 \\
			Employment status & 2.6 \\
			\bottomrule
		\end{tabular}
	\end{table}
\end{center}

\paragraph{Limitations.}
Complete-case analysis assumes that the excluded observations are not systematically different in ways that affect the outcomes conditional on included covariates. If missingness correlates with unobserved factors (MNAR), estimates may be biased. As a safeguard, we (i) report sample sizes next to each table, (ii) conduct robustness checks on alternative thresholds/definitions (which use overlapping but not identical subsets), and (iii) confirm that headline results are consistent across paired tests (McNemar/sign), proportion tests, and distributional comparisons.

\section{Binary vs.\ Ordinal Use of Multi-Point Scales}\label{app:binary-ordinal}

\paragraph{Why binarizing can reduce information and power.}
Collapsing a $K$-point ordinal scale (e.g., 1--5) into a binary indicator (e.g., top-box=1, else=0) discards the \emph{ordering and magnitude} among non-top categories. Statistically, this:
(i) reduces variance explained by any predictor (less signal; more residual variance),
(ii) attenuates effect sizes (coarser measurement induces misclassification around the cut), and
(iii) can move you from tests that use \emph{all} paired information (e.g., signed magnitudes/ranks) to those using \emph{only} signs or discordant pairs.
A classic benchmark: if a continuous/normal measure $X$ is dichotomized at its median, the correlation with the original is $\sqrt{2/\pi}\approx0.798$, implying $\approx36\%$ loss in variance explained and requiring $\sim1/0.798^2\approx1.57\times$ the sample size for the same power. Cutting at an \emph{extreme} (e.g., top-box) usually loses \emph{more} than this median split, because fewer observations carry signal.

\paragraph{How much power do we give up in practice?}
Relative efficiency (ARE) results provide a useful guide for paired tests under roughly symmetric noise:
\begin{itemize}
	\item Paired \textit{sign} test (uses only direction) vs.\ paired $t$ (uses magnitudes): $\text{ARE}\approx 2/\pi \approx 0.637$. 
	\emph{Implication:} to match the paired $t$'s power, a sign-based procedure needs $\approx 1/0.637 \approx 1.57\times$ more observations.
	\item Wilcoxon signed-rank (uses ranks/magnitudes without assuming interval spacing) vs.\ paired $t$: $\text{ARE}\approx 0.955$ under normal---nearly no power loss while keeping ordinal robustness.
\end{itemize}
Binarizing a 5-point outcome and testing proportions (or using McNemar/sign on paired binaries) moves you closer to the ``sign-only'' end of this spectrum, whereas using the \emph{full scale} with signed-rank or paired $t$ retains substantially more information.

\paragraph{Concrete sense of scale in our data.}
For H1, mapping the 5-point accuracy items to scores 1--5, the paired mean is $3.86$ at work vs.\ $3.08$ in personal life; the paired mean difference is $0.79$ (on a 1--5 scale), with standardized effect $d_z\approx0.81$ (based on the SD of within-person differences). With an effect of this size, both the binary and ordinal tests are highly powered in our sample. But in smaller samples or with subtler shifts (e.g., movement from ``Somewhat'' to ``Very'' without reaching ``Extremely''), the ordinal tests (paired $t$ or signed-rank) will \emph{detect} differences that a top-box binary may miss.

\paragraph{Why we still report a top-box (binary) view.}
Top-box indicators remain decision-relevant (e.g., ``what share insists on \emph{very high} accuracy?''), easy to interpret, and robust to individual scale-use idiosyncrasies. They also map cleanly to policy thresholds. That said, they are best seen as \emph{complements} to ordinal analyses, not substitutes.

\paragraph{Recommended practice (and what we do).}
We (i) report binary \emph{and} ordinal evidence side-by-side: paired McNemar/sign and proportion tests for the top-box contrast, plus distributional/ordinal tests (KS on full distributions, and---if desired---Wilcoxon signed-rank or paired $t$ on 1--5 scores); 
(ii) use ordinal or ordered-response models (e.g., ordered logit/probit) for determinants when the outcome is inherently graded; and 
(iii) reserve top-box binaries for interpretability and threshold-salience, noting their likely power cost (often equivalent to needing $\sim$1.5--2$\times$ the sample size compared to using the full 5-point scale, depending on where the cut is and the underlying distribution).

\section{Other Limitations in the Analysis}

\subsection{Scope of ``work'' and ``personal''.}
We treat work and personal life as broad categories, but stakes are heterogeneous within each domain. 
The consequences of an AI error for a surgeon or air-traffic controller are not comparable to those for a knowledge worker drafting internal communications; 
likewise, ``personal'' spans higher-stakes activities (e.g., personal finance, health queries) and low-stakes ones (e.g., entertainment playlists). 
Our average contrasts should therefore be interpreted as \emph{aggregate} differences across mixed task portfolios rather than task-invariant preferences. 
Future work should collect \emph{task-level} usage and perceived risk, enabling stratified estimates (by occupation, sector, and task stakes), vignette-based elicitations of acceptable error, and hierarchical models that separate within-person task heterogeneity from between-person differences.

\subsection{Interpretation of ``accuracy''.}
Respondents may interpret ``accuracy'' differently (e.g., factual correctness, logical coherence, absence of bias, or relevance to their query). 
Our items use the generic term, so reported preferences may conflate these facets. 
Future work should disaggregate accuracy into measurable subcomponents and validate with task-grounded exemplars to minimize construct ambiguity.

\subsection{Mechanisms remain black-boxed.}
While we document a robust gap in accuracy standards between work and personal contexts, we do not identify the causal \emph{mechanisms} behind this gap. 
Several explanations are plausible---role-based accountability and externalities at work \citep{ParasuramanRiley1997,LeeSee2004,Skitka2000}, 
perceived risk and loss aversion in professional settings, norms and incentives tied to job performance, and differential tolerance for experimentation in personal life \citep{Dietvorst2015,Logg2019}. 
A qualitative follow-up (semi-structured interviews or think-aloud protocols) on a purposive subsample and task-specific vignettes would help uncover how users interpret ``accuracy'', when they escalate verification, and why they set different thresholds across contexts.

\bibliographystyle{plainnat}
\bibliography{references}

\end{document}